\documentclass[11pt]{article}
\usepackage{coling2020}
\usepackage{times}
\usepackage{url}
\usepackage{multirow}

\usepackage{graphicx}
\usepackage{amsfonts}
\usepackage{amsmath}
\usepackage{subcaption}
\usepackage{pdflscape}
\usepackage{afterpage}
\usepackage{latexsym}
\usepackage{makecell}
\usepackage{todonotes}
\usepackage[toc,page]{appendix}

\usepackage{amssymb}

\colingfinalcopy 

\title{LAVA: Latent Action Spaces via Variational Auto-encoding\\for Dialogue Policy Optimization}

\author{Nurul Lubis, Christian Geishauser, Michael Heck, Hsien-chin Lin,\\\textbf{Marco Moresi, Carel van Niekerk and Milica Ga\v{s}i\'{c}}
 \\
  Heinrich Heine University D{\"u}sseldorf, Germany \\
  \small{\tt \{lubis,geishaus,heckmi,linh,moresi,niekerk,gasic\}@hhu.de} \\}

\date{}

\begin{document}
\maketitle
\begin{abstract}

Reinforcement learning (RL) can enable task-oriented dialogue systems to steer the conversation towards successful task completion. In an end-to-end setting, a response can be constructed in a word-level sequential decision making process with the entire system vocabulary as action space. Policies trained in such a fashion do not require expert-defined action spaces, but they have to deal with large action spaces and long trajectories, making RL impractical. Using the latent space of a variational model as action space alleviates this problem. However, current approaches use an uninformed prior for training and optimize the latent distribution solely on the context. It is therefore unclear whether the latent representation truly encodes the characteristics of different actions. In this paper, we explore three ways of leveraging an auxiliary task to shape the latent variable distribution: via pre-training, to obtain an informed prior, and via multitask learning. We choose response auto-encoding as the auxiliary task, as this captures the generative factors of dialogue responses while requiring low computational cost and neither additional data nor labels. Our approach yields a more action-characterized latent representations which support end-to-end dialogue policy optimization and achieves state-of-the-art success rates. These results warrant a more wide-spread use of RL in end-to-end dialogue models.
\end{abstract}

\blfootnote{
   %
   %
   %
   %
   %
   %
   \hspace{-0.65cm}  
   This work is licensed under a Creative Commons
   Attribution 4.0 International License.
   License details:
   \url{http://creativecommons.org/licenses/by/4.0/}.
}

\section{Introduction}
\label{intro}

With the rise of personal assistants, task-oriented dialogue systems have received a surge in popularity and acceptance. Task-oriented dialogue systems are characterized by a user goal which motivates the interaction, e.g. booking a hotel, searching for a restaurant, or calling a taxi. The dialogue agent is considered successful if it is able to fulfill the user goal by the end of the interaction. Traditionally, a dialogue system is built using the divide and conquer approach, resulting in multiple modules that together form a dialogue system pipeline: a natural language understanding (NLU) module, a dialogue state tracker (DST), a dialogue policy, and a natural language generation (NLG) module. Each module has well-defined input and output, and can be trained using machine learning~\cite{ygkm10,thyo10,will06,hety13} provided that adequately labeled data is available. However, there is a loss of information between the modules. The availability of powerful deep learning methods has recently led to a surge in end-to-end training approaches~\cite{bowe16,wvmg17,madotto2018mem2seq}, which aim to map user utterances directly to responses in a sequence-to-sequence fashion.

Two fundamental properties of task-oriented systems are the ability to remember everything that is important from the conversation so far -- \emph{tracking}, and the ability to produce a response that steers the conversation towards successful task completion -- \emph{planning}. Within the modular approaches the role of tracking is taken by the DST, optimized using supervised learning~(SL), while the role of planning is taken by the dialogue policy, optimized via reinforcement learning~(RL). Unlike their modular counterparts, end-to-end systems typically only deploy SL, which relies on language modeling techniques that directly optimize the likelihood of the data under the model parameters, neglecting planning altogether. This line of research has hugely benefited from large pre-trained transformer-based models such as BERT and GPT-2~\cite{devlin2019bert,radford2019language,hosseini2020simple,peng2020soloist}. 

Only few works in end-to-end task-oriented dialogue systems deploy RL~\cite{mehri2019structured,zhao2019rethinking}. Word-level RL views each word of the entire system vocabulary as an action in a sequential decision making process. This blows up the action space size and the trajectory length, hindering effective learning and optimal convergence. The challenge of credit assignment and reward signal propagation is further compounded by the typically sparse rewards in dialogue. Last but not least, simultaneously optimizing language coherence and decision making within one model can lead to divergence. Thus, effectively incorporating RL in the end-to-end setting remains a challenge.

In the recently proposed latent action reinforcement learning (LaRL), a latent space between the context encoder and the response decoder serves as action space of the dialogue agent~\cite{zhao2019rethinking}. Decoding responses conditioned on the latent variable has the benefit of decoupling action selection and language generation, as well as shortening the dialogue trajectory, leading to improved performance. However, this approach optimizes the latent space using an uninformed prior without taking into consideration the actual distribution of the responses. Furthermore, the latent space model is conditioned only on the context. Therefore it is unclear whether the latent variables truly encode the characteristics of different dialogue actions, or whether it encodes the dialogue context instead. Because RL optimizes action selection and planning, it is important that it is performed on an action-characterized space.

In this paper, we propose an unsupervised approach for optimizing the latent action representation for end-to-end dialogue policy optimization with RL. Our contributions are as follows:
\begin{itemize}
    \item 
    We propose to optimize latent representations to be action-characterized. Action-characterized representations encode similar actions close to each other and allow interpolation of actions. This leads to a more practical and effective end-to-end RL. 
    \item 
    We explore three ways of leveraging an auxiliary task to shape the latent variable distribution; via pre-training, to obtain an informed prior, and via multitask learning. As auxiliary task, we choose response auto-encoding, as this captures generative factors of the dialogue responses. Unlike contemporary transformer-based approaches, this requires no additional data and has low computational cost. Our analysis shows that the learned latent representations encode action-relevant information. 
    \item 
    We show that our approach achieves state-of-the-art match and success rates on the multi-domain MultiWoZ 2.0\footnote{The codebase is accessible at: \texttt{https://gitlab.cs.uni-duesseldorf.de/general/dsml/lava-public}. We also achieve state-of-the-art results on MultiWoZ 2.1}~\cite{budzianowski2018large}. 
\end{itemize}

This work acts as a proof of concept that we can induce an action-characterized latent space in an unsupervised manner to facilitate more practical and effective RL. 
The overall performance could likely be improved further by using more sophisticated encoding and decoding models, but this goes beyond the scope of this work. 
The results obtained here already warrant a more wide-spread use of RL in end-to-end dialogue models. 

\section{Related Work}
Research in end-to-end task-oriented systems is largely inspired by the success of sequence-to-sequence modeling for chat-oriented systems~\cite{serban2016building}. Representation learning has been shown to be useful for end-to-end systems, allowing more effective information extraction from the input. A common method leverages pre-training objectives that are inspired by natural language processing tasks, e.g. next-utterance retrieval~\cite{lowe2016evaluation} and generation~\cite{vinyals2015neural}. Naturally, this requires sufficient amounts of additional data, and often labels. The choice of the pre-training objective has been demonstrated to highly influence generalizability of the learned representation~\cite{mehri-etal-2019-pretraining}.

More recently, researchers have also investigated representation learning towards a better modeling of dialogue response. For example,~\newcite{zhao2020learning} have investigated the use of language modeling tasks such as masking and sequence ordering for response generation. With variational models, latent space that spans across domains can be induced using dialog context-response pairs as well as a set of response-dialog act pairs~\cite{zhao2018zero}. Similarly, dialogue context preceding and succeeding a response can be used in a skip-thought fashion to train response embeddings ~\cite{zhao2018unsupervised}. It has been reported that such representations allows few-shot domain adaptation using only raw dialog data ~\cite{shalyminov2019few}. State labels and their transitions have also been utilized for learning action representations \cite{huang2019mala}. Performing RL on top of the learned latent variable space has been shown to lead to a better policy compared to word-level RL, due to the condensed representation and shorter trajectory \cite{zhao2019rethinking}. While improvement on metrics such as task success and entity recognition are reported, lack of interpretability and controllability remains a major challenge in this family of models.

\section{Preliminaries}
\label{sec:preliminaries}
In an end-to-end framework, dialogue policy optimization with RL typically consists of two steps: SL and policy gradient RL. In the SL step, the model learns to generate a response $x$ based on some dialogue context $c$, updating its parameters $\theta$ to maximize the log likelihood of the data,

\begin{equation}
	\label{eq:loss_sl}
	L_\textup{SL}(\theta) = \mathbb{E}_{x, c}[\log p_\theta(x|c)].
\end{equation}

\noindent Subsequently, the RL step updates the model parameter w.r.t. the task-specific goal, reflected as a reward.
In a dialogue with $T$ steps, for a specific time-step $t$, immediate reward $r_t$, and discount factor $\gamma \in [0, 1]$, the discounted return is defined as $R_t = \sum_{k=t}^{T} \gamma^{k-t}r_{k}$. The model tries to maximize the expected return from the first time-step onwards, written as $J(\theta)=\mathbb{E_\theta}[\sum_{t=0}^{T}\gamma^t r_t]$.

In word-level RL, every output word is treated as an action step, yielding the following policy gradient:
\begin{equation}
	\nabla_\theta J(\theta)=\mathbb{E}_\theta[\sum_{t=0}^{T}\sum_{j=0}^{U_t}R_{tj}\nabla_\theta \log p_\theta(w_{tj}|w_{<tj}, c_t)]
\end{equation}

\noindent where $T$ is the total number of turns in the dialogue, $U_t$ is the total number of tokens in the response at turn $t$ and $j$ is the index of each token $w$. $R_{tj}$ denotes the discounted return of the $j$-th token at turn $t$. In this policy gradient form, the action space is the vocabulary size of the system $|V|$, and the trajectory length is $\sum_{t=0}^{t=T}U_t$, making RL in this space extremely challenging.

The introduction of a latent variable $z$ allows us to factorize the conditional distribution into $p(x|c) = p(x|z)p(z|c)$. By treating the latent space $z$ as the action space, the action space size and trajectory length are reduced \cite{zhao2019rethinking}. For a dialogue with $T$ turns, policy gradient is now given by
\begin{equation}
	\nabla_\theta J(\theta)=\mathbb{E}_\theta[\sum_{t=0}^{T}R_t\nabla_\theta\log p_\theta(z_t|c_t)],
	\label{eq:reinforce}
\end{equation}



\noindent where $R_{t}$ denotes the discounted return at turn $t$. \newcite{zhao2019rethinking} have examined two types of latent variables; categorical and continuous. The categorical latent variable takes form as $M$ independent $K$-way categorical random variables, while the continuous one is modeled as $M$ dimensional multivariate Gaussian distribution with a diagonal covariance matrix. The latent variable distribution is learned during the SL step using stochastic variational inference by maximizing the evidence lowerbound (ELBO) -- the lowerbound on the data log likelihood, 
\begin{equation}
	\label{eq:full_elbo}
 	L_\textup{full}(\theta)=\mathbb{E}_{q_\theta(z|x, c)}[\log p_\theta(x|z)] -\textup{D}_\textup{KL}[q_\theta(z|x, c)||p_\theta(z|c)].
\end{equation}

To combat exposure bias,~\newcite{zhao2019rethinking} introduced a ``lite'' version of ELBO by assuming the posterior $q_\theta(z|x, c)$ to be the same as the encoder $p_\theta(z|c)$. This eliminates the second term of the ELBO objective. To counter overfitting, the posterior is regularized with some weight $\beta$ to be similar to certain priors, in this case a uniform distribution for categorical latent variables, or a normal distribution for continuous ones. The ``lite'' ELBO objective has been shown to outperform the full ELBO objective, and is written as:
%
\begin{equation}
	\label{eq:lite_elbo}
	L_\textup{lite}(\theta)=\mathbb{E}_{p_\theta(z|c)}[\log p_\theta(x|z)] - \beta \textup{D}_\textup{KL}[p_\theta(z|c)||p(z)].
\end{equation}



\section{Latent Action Space via Auxiliary Task}
\label{sec:vaux}

\begin{figure}
	\centering
	\includegraphics[width=0.8\linewidth]{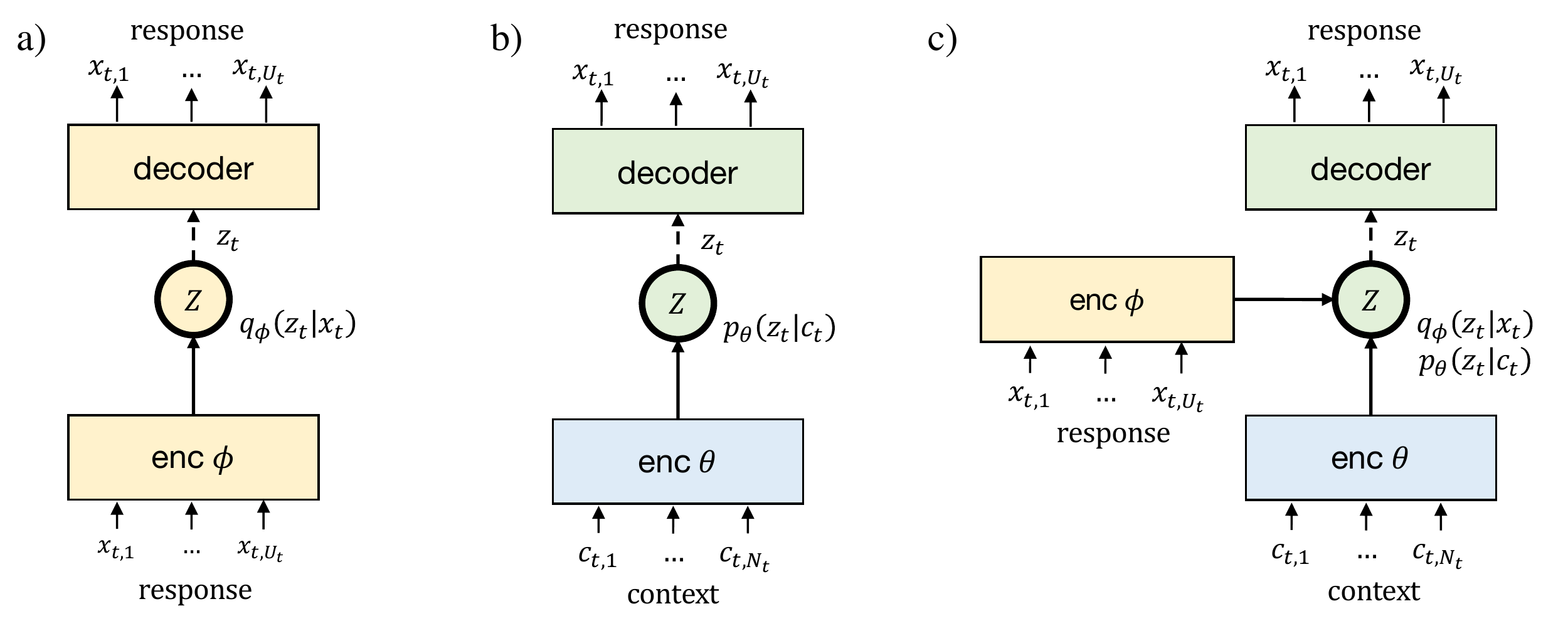}
	\caption{We aim to train an action-characterized distribution of latent variables $Z$ to support RL in an end-to-end setting. a) VAE pre-training, b) new encoder is initialized and connected to the pre-trained $Z$ and VAE decoder. Overall fine-tuning optimizes the entire network (LAVA\_ptA), while selective fine-tuning optimizes only the new encoder (LAVA\_ptS), c) new encoder is initialized and used in tandem with VAE encoder to obtain informative prior (LAVA\_kl). The same architecture is also used for multitask learning (LAVA\_mt), where we optimize for both tasks at the same time from scratch.}
	\label{fig:lava}
\end{figure}

As evidenced in Equation (\ref{eq:reinforce}), to facilitate an effective learning it is important that the policy is trained using latent variables $z$ that meaningfully represent actions. However, because existing methods model the distribution of $z$ w.r.t. the context $c$ (Equation (\ref{eq:lite_elbo})), we suspect that the latent variable is rather a representation of the state space instead of the action space. Furthermore, the distribution is regularized using an uninformed prior and without taking into account the actual distributions regarding dialogue responses for a given context. To induce a more action-like latent representations, we train the model on an auxiliary task that requires knowledge of responses to perform. We chose response auto-encoding (AE) as the auxiliary task using the variational auto-encoding (VAE) model. That is, given a response $x$ we train the model to reconstruct the response via a latent space between the encoder and decoder (Figure~\ref{fig:lava}a). With an uninformed prior $p(z)$, the pre-training objective for a set of parameters $\phi$ is:
\begin{equation}
\label{eq:vae}
L_\textup{vae}(\phi)=\mathbb{E}_{q_\phi(z|x)}[\log p_\phi(x|z)] -\textup{D}_\textup{KL}[q_\phi(z|x)||p(z)]. 
\end{equation}

\noindent VAE models have been shown to be able to capture generative aspects of the samples they are trained on, resulting in good interpolation between latent variables \cite{kingma2014auto,bowman2016generating}. By training a VAE on dialogue responses, we aim to capture generative aspects of responses such as intent and domain information in an unsupervised manner.

We propose to utilize the VAE latent representations to condition dialogue systems to map encoded dialogue states to latent actions, instead of learning latent representations of the dialogue states. We call this approach \textbf{LAVA} (Latent Action via VAE). We explore three ways of leveraging the AE task to induce action-characterized latent variable distributions: as pre-training, as informed prior, and in a multitask learning fashion -- pictured in Figure~\ref{fig:lava}. Note that it is possible to swap the AE task with other tasks that target representation learning on the dialogue responses. In this work we utilize simple recurrent models as encoder and decoder to highlight the role of the latent dialogue action space. This allows us to pin any observed improvements on the latent space and dialogue policy. Other parts of the end-to-end dialogue system framework, such as encoding and decoding, are not within the scope of this work. 

\subsection{Auxiliary Task as Pre-training} 
\label{ssec:aux_pt}
The first method utilizes the auxiliary AE task to pre-train the latent representation and decoder (Figure~\ref{fig:lava}b). Since the vocabularies of user and system turns vastly differ, a new dialogue system encoder for response generation (RG) is initialized and the VAE encoder is discarded. The new encoder is connected to the latent space and the decoder of the VAE. We experiment with two kinds of fine-tuning scheme: overall (LAVA\_ptA) and selective (LAVA\_ptS). With LAVA\_ptA, we update all parameters of the network during fine-tuning. On the other hand, LAVA\_ptS blocks the gradient propagation to the latent space and the decoder and exclusively trains the encoder. This is equivalent to utilizing the generation modules from the VAE to serve as the NLG module, encouraging the model to focus on encoding the dialogue state (similar to a state tracker) and mapping it to a corresponding latent action (similar to a policy) during RG training. The loss function for both LAVA\_ptA and LAVA\_ptS is the ``lite'' ELBO ebjective (Equation \ref{eq:lite_elbo}).

\subsection{Auxiliary Task as an Informative Prior}
Secondly, we explicitly train the RG latent variable distribution to be close to that of the VAE. We call this set up LAVA\_kl. As before, we start with a pre-trained VAE and a newly initialized RG encoder. We exploit the pre-trained latent representation in a novel way; the VAE encoder is not discarded and instead used in tandem to obtain an informed prior of the target response, pictured in Figure~\ref{fig:lava}c. We use the latent distribution conditioned on the target $q_\phi(z|x)$ in the KL term penalty, replacing the uninformed prior used in previous works. This grounds the RG latent variable distribution to that of the VAE by penalizing divergence, while still optimizing it to fit the dialogue contexts. All parameters except the VAE encoder are further optimized with the following ELBO:
\begin{equation}
\label{eq:lite_elbo_kl}
L_\textup{LAVA\_kl}(\theta)=\mathbb{E}_{p_{\theta}(z|c)}[\log p_\theta(x|z)] - \beta \textup{D}_\textup{KL}[p_{\theta}(z|c)||q_{\phi}(z|x)].
\end{equation}

\subsection{Multitask Training between Main and Auxiliary Tasks} 

Lastly, we train a model to solve the RG and AE tasks in a multitask fashion, where RG is considered as the main task and AE as the auxiliary task. Multitask learning aims to improve learning efficacy by having a model solve multiple tasks at once, exploiting similarities across tasks~\cite{caruana1997multitask}. Recent works have shown that dialogue system tasks also benefit from multitask learning~\cite{rastogi2018multi,zhu2019multi}. RG and AE tasks are similar in that both aim to generate dialogue responses $x$, but they differ in the context they consider, also called the many-to-one multitask setting \cite{luong2015multi}. RG attempts to generate the target response $x$ given a dialogue context $c$, and AE tries to perform reconstruction given a response $x$. The two tasks share the latent space and decoder with a set of parameters $\omega$ but with separate encoders for RG and AE, with parameters $\theta$ and $\phi$, respectively  (Figure~\ref{fig:lava}c). The aim is that the latent representation encodes more action-characterized features, since these are the common information required to fulfill both tasks. The two tasks are trained in an alternate fashion with an A:B ratio, i.e. for every A iterations of the main task, we train with the auxiliary task for B iterations. Unlike the previous methods, in multitask learning we start with a newly initialized model without pre-training. Each encoder receives an update only from its corresponding task, while the latent representation and decoder are trained on both tasks. The ELBO objectives are
\begin{equation}
	L_\textup{LAVA\_mt}^\textup{RG}(\omega, \theta) = \mathbb{E}_{p_{\theta}(z|c)}[\log p_\omega(x|z)] - \beta \textup{D}_\textup{KL}[p_{\theta}(z|c)||p(z)],
\end{equation}
\begin{equation}
	L_\textup{LAVA\_mt}^\textup{AE}(\omega, \phi) = \mathbb{E}_{q_{\phi}(z|x)}[\log p_\omega(x|z)] - \beta \textup{D}_\textup{KL}[q_{\phi}(z|x)||p(z)].
\end{equation}

\section{Experiment Setup}
\subsection{Corpus, Task, and Training Setup}

We use the MultiWOZ 2.0 corpus~\cite{budzianowski2018large} to test the performance of the models. MultiWOZ is a collection of conversations between humans in a Wizard-of-Oz fashion, where one person plays the role of a dialogue system and the other one a user. The user is tasked to find entities, e.g. a restaurant or a hotel, that fit certain criteria by interacting with the dialogue system. The corpus simulates a multi-domain task-oriented dialogue system interaction, i.e. multiple domains may occur in the same dialogue or even the same turn. The corpus is fully annotated with a total of 10438 dialogues in English, it is one of the most challenging and largest corpora of its kind. We use the training, validation, and test set partitions provided in the corpus, amounting to 8438 dialogues for training, and 1000 each for validation and testing. All numbers reported are based on evaluation on the test set.

We aim to train a latent action representation that is effective for optimizing dialogue policies with RL in an end-to-end setting. This goal is best reflected by completion of the underlying dialogue task, measured in dialogue-level match and success rates. Match rate computes whether the criteria informed by the user (informable slots) are matched by the system, and success rate computes whether information requested by the user (requestable slots) are provided by the system. Match is a pre-requisite for a successful dialogue. For a long time the research in dialogue policy has only looked at success rates and user satisfaction~\cite{lees12,ultes2017domain}, but as the line between policy and NLG becomes blurred we see the introduction of metrics such as BLEU and perplexity. However, these have been labeled early on to be potentially misleading as they correlate poorly with human judgement~\cite{stent2005evaluating,liu2016not}. Although we also report BLEU score for completeness, note that our methods are not targeted at improving BLEU. 

We examine two tasks: 1) ``dialogue context-to-response generation,'' that is to generate the next dialogue response given dialogue history, as well as oracle dialogue state and database pointer from the corpus. The dialogue state is a binary vector representation of the user goal as inferred from the beginning of the dialogue up to the current turn. On the other hand, the database pointer is a binary vector representation of the entity matching the criteria in the dialogue state. 2) ``End-to-end modeling" which takes only dialogue history for response generation. Works in end-to-end modeling typically utilize intermediate models in the pipeline to predict labels such as dialogue state and database pointer. In this work we utilize the latent action in an end-to-end fashion without the use of any intermediate labels, encouraging the model to fully exploit the latent variables. All experiments are conducted on delexicalized dialogues, where occurences of slot values are replaced with their corresponding slot tokens, for example ``in southern part of town'' becomes ``in \texttt{[value\_area]} part of town.'' 



Our training consists of two steps: techniques presented in Section \ref{sec:vaux}, followed with RL using REINFORCE. For a fair comparison with existing works, we adopt the novel RL setup proposed by \newcite{zhao2019rethinking}: 1) For each RL episode, sample a dialogue from the corpus. 2) Run the model to generate a response for each system turn. However, the next user turn in the dialogue is not altered and simply retrieved from the corpus. 3) Compute success rate of the dialogue based on system response and use this as reward signal to compute policy gradient and update model parameters (Equation~\ref{eq:reinforce}).

\subsection{Model}
Our primary focus are the latent representations induced by the proposed methods, their effect on reinforcement learning and the final performance of the model. To highlight the role of the latent dialogue action space, we use simple recurrent models as encoder and decoder for both the VAE and the dialogue system. We limit the model vocabulary to the most frequent 1000 tokens. We truncate the dialogue history to the last 2 turns for the context-to-response task, and the last 4 turns for the end-to-end modeling task. For both AE and RG tasks, the encoder is a GRU-LSTM~\cite{cho2014learning} with attention and size 300 which outputs a vector with size 600. We tested both categorical and continuous latent variables. The categorical latent space is 10 independent 20-way categorical variables ($M=10$, $K=20$) and the continuous space is set to size $M=200$. In the categorical case, the decoder is of size 150 with attention, and in the continuous case the decoder is of size 300. In choosing the hyperparameters, we follow the experimental set up reported by~\newcite{zhao2019rethinking}. We tested a few different set-ups, for example by varying the size of the latent space or the network, but found that the reported settings give the most optimal performance. One exception is the weight $\beta$ for the KL term, which we set to 0.01 for all models other than LAVA\_kl, which performs best with 0.1. Multitask training ratio is set to 10:1. 

Unlike transformer-based architectures, the training of our models are computationally light and fast. One training takes between 1-3 hours using a single RTX 2080 GPU. While typical SL training requires 80-85 epochs with batch size of 128, LAVA\_kl converges in under 20 epochs. LAVA\_ptA, LAVA\_ptS, and LAVA\_mt convergence varies at around 20-50 epochs. LAVA models likely benefit from the pre-trained VAE models and is therefore able to converge faster.


\section{Experimental Results}
\subsection{Context-to-Response Generation}

\begin{table}
    \small
    \centering 
		\begin{tabular}{|l|rr|r|rr|r|}
			\hline
			\multirow{2}{*}{Model} & \multicolumn{3}{c|}{SL}                 & \multicolumn{3}{c|}{+RL}  \\ \cline{2-7}
			&                      match & succ. & BLEU  & match & succ. & BLEU \\ \hline
			
			LiteAttnCat*  & 65.77 &  57.26 &   0.18  & 83.68   & 78.18     & 0.12  \\
			     
			LAVA\_ptA\_cat   &   70.57 &  58.56 &   0.18 & 85.09   & 77.08     & 0.12    \\ 
			     
			LAVA\_ptS\_cat &  64.56 &  54.65 &   0.19 &  83.48   & 79.87     & 0.12  \\ 
			LAVA\_kl\_cat   &   71.97 &  57.96 &   0.18 &  \textbf{85.59}   & \textbf{83.38}     & 0.12   \\ 
			LAVA\_mt\_cat        &     60.46 &  51.05 &   0.18  & 84.88   & 80.98     & 0.10 \\ 
			\hline \hline
			LiteGauss* &    68.27 &  57.06 &   0.19 & 75.88   & 66.47     & 0.14  \\ 
			LAVA\_ptA\_gauss      &   64.46 &  54.55 &   0.19 & 77.78   & 62.26     & 0.15  \\ 
			LAVA\_ptS\_gauss &     67.37 &  53.85 &   0.18 &  77.78   & 61.76     & 0.13   \\ 
			LAVA\_kl\_gauss & 68.87 &  58.66 &   0.19  & 79.28 &   64.96  &  0.09  \\ 
			LAVA\_mt\_gauss         &   59.42 &  49.33 &   0.19 &  82.78   & 70.37     & 0.14   \\ \hline\hline 
			Seq2Seq &   58.66 &  52.25 &   0.20  & 81.58   & 75.18     & 0.15    \\ 
			SFN \cite{mehri2019structured} & 65.80 & 51.30  & 0.17 & 82.70 & 72.10 & 0.16   \\
			LiteAttnCat \cite{zhao2019rethinking}   &67.97 &57.36 & 0.19 & 82.80 & 79.20 & 0.12  \\
			\hline 
			
			
		\end{tabular}
		\captionof{table}{Best performance of our proposed methods in comparison with *reproduced and reported baselines methods that employ RL. Our best model LAVA\_kl\_cat surpasses the baselines in both match and success rates. }
    \label{tab:proposed_methods}
\end{table}

Table~\ref{tab:proposed_methods} presents the performance of our models in comparison with 1) sequence-to-sequence (Seq2Seq) and structured fusion network (SFN) as baseline models that do not employ latent variables and 2) LaRL models, LiteAttnCat and LiteGauss, which we reproduced using the public code (marked with *) \cite{zhao2019rethinking}. The Seq2Seq model shows best BLEU score compared to any of our models, however its match and success rate are consistently lower than our categorical models. This is not surprising, as Seq2Seq is optimized only to maximize the likelihood of the data. With the categorical latent variable, training a model on top of the pre-trained VAE, either in a selective manner or not, improves dialog-level performance. Using the VAE as informed prior gives us improvements on both match and success rates while maintaining the BLEU score, resulting in our best performing model. Although we find that continuous latent space is not as effective for RL, the proposed multitask training still surpasses the LiteGauss baseline and gives the best performance when Gaussian latent space is utilized.

\begin{table}[t]
\small
    \centering 
    \begin{tabular}{|l|c|c|c||c|c|}
        \hline
        Model              & Match & Success & BLEU & Transformer & RL\\ \hline
        Human & 90.40 & 82.28 & - & - & -  \\ \hline \hline
        SimpleTOD \cite{hosseini2020simple} & 88.90 & 67.10   & 0.16 & $\checkmark$ & -\\ \hline
        ARDM \cite{wu2019alternating} & 87.40 & 72.80   & 0.20 & $\checkmark$ & - \\ \hline
        DAMD \cite{zhang2019task} & 89.20 & 77.90   & 0.18 & $\checkmark$ & -\\ \hline
        SOLOIST \cite{peng2020soloist} & 89.60 & 79.30   & 0.18 & $\checkmark$ & -\\ \hline
        MarCo \cite{wang2020multi} & 92.30 & 78.60   & 0.20 & $\checkmark$ & -\\ \hline
        HDNO \cite{wang2020modelling} & 96.40 & 84.70   & 0.18 & - & $\checkmark$ \\ \hline \hline
        LAVA\_kl\_cat + RL (ours) & \textbf{97.50} & \textbf{94.80}   & 0.12 & - & $\checkmark$ \\ \hline 
    \end{tabular}
    \caption{Comparison of our best performing model with existing works on the same task. For a fair comparison we adjust the performance of our best model by recalculating the match and success rates using a modified script released in the MultiWoZ repository$^2$. We also note whether the approaches employ a transformer-based architecture and RL.}
    \label{tab:sota}
\end{table}

We also compare our best performing model with existing works tackling the same task. Unlike the baseline models in Table \ref{tab:proposed_methods}, these works are evaluated with a new evaluation script recently published in the official MultiWoZ repository. The new evaluation script differs to the original one in the treatment of one specific case in the train domain, where a train matching user requirement is found but the user does not request the train ID. The original script underestimates the model performance as the train ID is checked regardless, while the new evaluation script do not check this further. For a fair comparison with relevant state-of-the-art models, we re-compute the match and success rates of the model using the new evaluation script. The numbers are reported in Table~\ref{tab:sota}.

We reach state-of-the-art inform and success rates, surpassing that of existing works which tackle the same task and even human performance on the test set at 82.28\%. This result demonstrates the advantage of RL in the end-to-end setting. Optimizing for reward allows us to reach a higher success rate compared to solely focusing on accurately generating the target response. Our reward definition only takes into account the dialogue success rate, however because match is a prerequisite of dialogue success we are able to harmoniously optimize both match and success rates during RL. It is also interesting to note that our best model has higher match to success ratio compared to existing works, i.e. , we are able to achieve success on most dialogue where match occurs, while existing works fail more often in providing user with the information they require even when match already occurs. Moreover, unlike state-of-the-art models, our end-to-end setting uses simple encoder-decoder models without explicit dialogue state tracking, and therefore it is evident that the improvements are contributed by the latent action space. Combining our methods with more powerful models would be straightforward and we expect it to further improve the performance. 

Notwithstanding, with regards to the very high performance, it is important to note the limitation of the current evaluation set up, most importantly that the dialogue trajectory is only estimated, since the user turn is obtained from data regardless of system response. To better gauge the performance in real dialogue with humans, user evaluation needs to be conducted in the future.

\subsection{End-to-End Generation}
We also tested our latent action in a fully end-to-end fashion, i.e. without using dialogue state labels and database pointers. Unlike existing works that train intermediate models for predicting labels such as dialogue state and action, our aim is to rely solely on the latent variable for forming a dialogue policy. We take our LAVA\_kl\_cat as pre-trained model, and further perform SL and RL exclusively with raw dialogue data in an end-to-end setting. 


We present the performance of our best model in comparison with existing works in Table~\ref{tab:e2e}, along with the types of labels they utilize in the pipeline. Performance of these models are also computed with the new evaluation script as in Table~\ref{tab:sota}. Consistent with previous task, our model is outperforming the other models in terms of success and inform rates. The result confirms that our model is able to optimize its dialogue policy by leveraging action-relevant information that is encoded in the latent variable, e.g. action type and domain, even in an extreme setting where no additional label is utilized in the pipeline.

\begin{table}
    \small \centering
    \begin{tabular}{|l|ccc|rr|r|}
    \hline
            \multirow{2}{*}{Model}& \multicolumn{3}{c|}{labels} & \multicolumn{2}{c|}{dialog-level} & \multicolumn{1}{c|}{turn-level}\\ \cline{2-7}
            &                   dialogue state & DB Search & action  &  match & succ. &BLEU  \\ \hline
  			DAMD \cite{zhang2019task} & gen & oracle & gen & 76.30 & 60.40 & 0.18  \\
  			SimpleTOD \cite{hosseini2020simple} & gen & - & gen & 84.40 & 70.10 & 0.15   \\
  			SOLOIST \cite{peng2020soloist} & gen & gen & - & 85.50 & 72.90 & 0.16  \\
 			\hline \hline
 			LAVA\_kl\_cat + RL & - & - & latent & \textbf{91.80} & \textbf{81.80} & 0.12 \\ 
 			\hline
    \end{tabular}
    \caption{Comparison to state-of-the-art models on end-to-end generation task. Labels can come from the corpus (oracle), prediction using supervised models (gen), or in our case from latent space (latent). Our model is able to perform well without any additional labels by solely leveraging the latent variable.}
    \label{tab:e2e}
    \vspace{-0.5cm}
\end{table}

\section{Latent Space Analysis}
\label{sec:analysis}

\subsection{Clustering Metrics and Projection}
\label{ssec:clustering_metrics}

We investigate whether the latent variables are grouped together according to true action or domain labels. One method is to quantify the quality of clusters that are formed in the latent space w.r.t. action and domain labels. We use the Calinski-Harabasz index, which measures the ratio of the sum of between-clusters dispersion and of inter-cluster dispersion for all clusters \cite{calinski1974dendrite}. A higher Calinski-Harabasz score relates to a model with better defined clusters\footnote{We note that extremely high scores could signal cluster overfitting. See Appendix~\ref{app:overfit} for more details.}.

Table~\ref{tab:clustering_metrics} compares the scores of the LAVA\_kl and LAVA\_mt methods with their corresponding reproduced LaRL baselines~\cite{zhao2019rethinking}. We observe that for all models domain labels are better clustered than action, which is expected because 1) the amount of unique domains is much smaller than that of actions, and 2) domain information is explicitly expressed at word level, while action is concerned with the intent of the utterance. Note that since domain information is part of the action label, efficient domain clustering is also beneficial for inferring actions. Performing RL on top of SL consistently improves the clustering scores. However, in some cases improvement on domain clusters comes at the cost of action cluster, e.g. categorical LaRL and LAVA\_kl\_gauss. We observe that good dialogue performance aligns with harmonious improvement of domain and action clusters, as well as a nice balance between the domain and action scores, as demonstrated by LAVA\_kl\_cat and LAVA\_mt\_gauss.




\begin{table}
	\centering \small
	\begin{tabular}{|l||r|r|r|r||r|r|r|r|}
		\hline
		\multirow{3}{*}{Model}& \multicolumn{4}{c||}{Categorical}                       & \multicolumn{4}{c|}{Gaussian}                    \\ \cline{2-9}
		& \multicolumn{2}{c|}{SL} & \multicolumn{2}{c||}{RL} & \multicolumn{2}{c|}{SL} & \multicolumn{2}{c|}{RL} \\ \cline{2-9}
		 & Domain & Action&Domain & Action&Domain & Action&Domain & Action \\ \hline \hline
		LaRL*    &  93.19 &  23.30 & 121.15 &   17.50 &         47.86  &           13.04&             71.49 &             13.25  \\    
		LAVA\_kl   & 104.92            & 25.28 & 158.00 &             41.75      & 106.51            & 20.00     &  128.51 &             19.13     \\
		LAVA\_mt &       25.37 &            6.64  & 247.92 &           16.85 &               66.54 &           16.91 &  80.09 &             18.50 \\ 
		\hline
	\end{tabular}
	\caption{Clustering metrics scores $\uparrow$. Our LAVA\_kl\_cat and LAVA\_mt\_gauss show harmonious improvement of domain and action clusters, as well as a nice balance between the domain and action scores.}
	\label{tab:clustering_metrics}
	\vspace{-0.5cm}
\end{table}

\begin{figure}[ht]
\centering
	\begin{subfigure}{.24\textwidth}
		\centering
		\includegraphics[width=\linewidth]{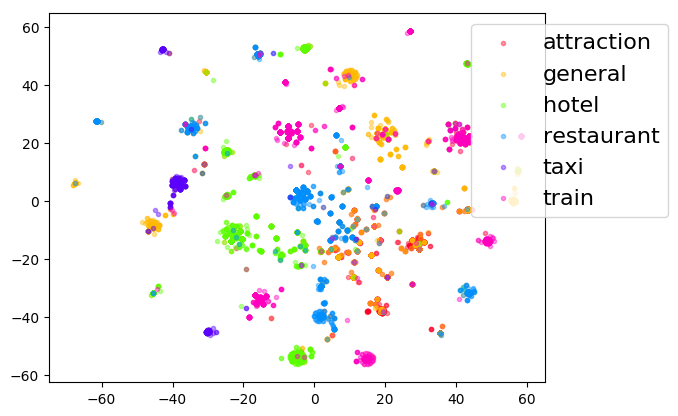}
		\caption{LAVA\_kl\_cat, domain}
	\end{subfigure}%
	\begin{subfigure}{.24\textwidth}
		\centering
		\includegraphics[width=\linewidth]{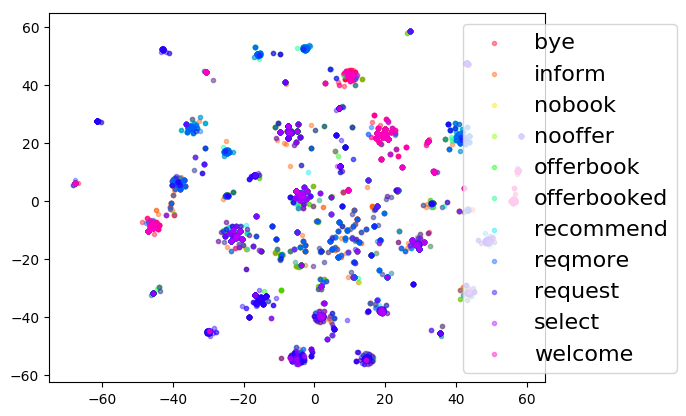}
		\caption{LAVA\_kl\_cat, action}
	\end{subfigure}%
	\begin{subfigure}{.25\textwidth}
  		\centering
  		\includegraphics[width=\linewidth]{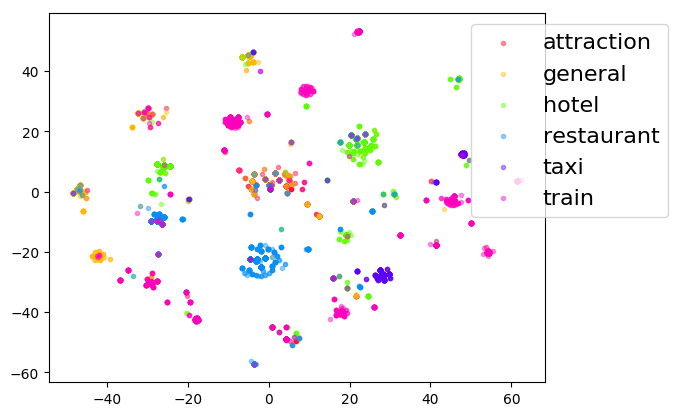}
  		\caption{LAVA\_kl\_cat+RL, domain}
  	\end{subfigure}%
  	\begin{subfigure}{.25\textwidth}
  		\centering
  		\includegraphics[width=\linewidth]{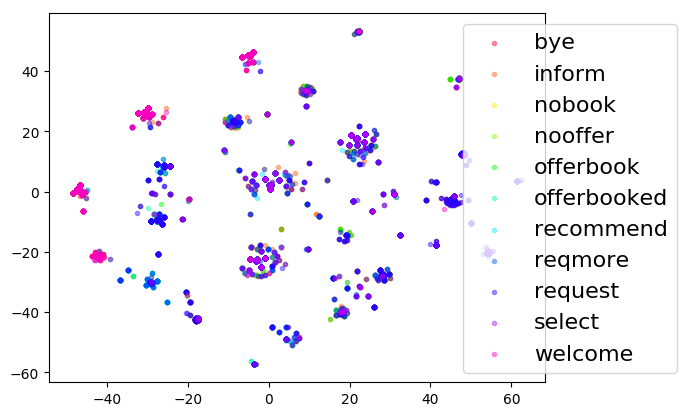}
  		\caption{LAVA\_kl\_cat+RL, action}
	\end{subfigure}
 	\label{fig:actz_cat_rl_cluster}
	\begin{subfigure}{.24\textwidth}
		\centering
		\includegraphics[width=\linewidth]{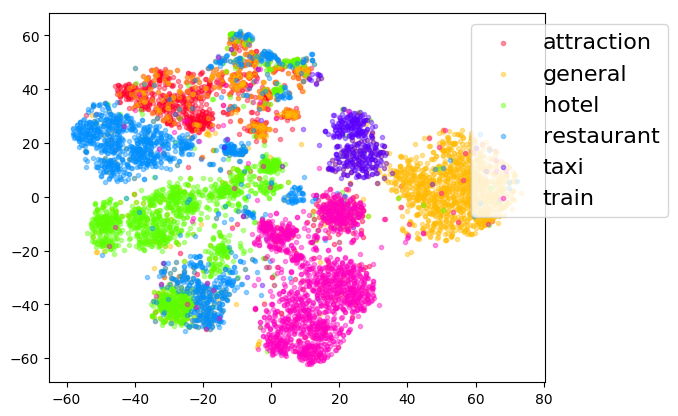}
		\caption{LiteAttnCat*, domain}
	\end{subfigure}%
	\begin{subfigure}{.24\textwidth}
		\centering
		\includegraphics[width=\linewidth]{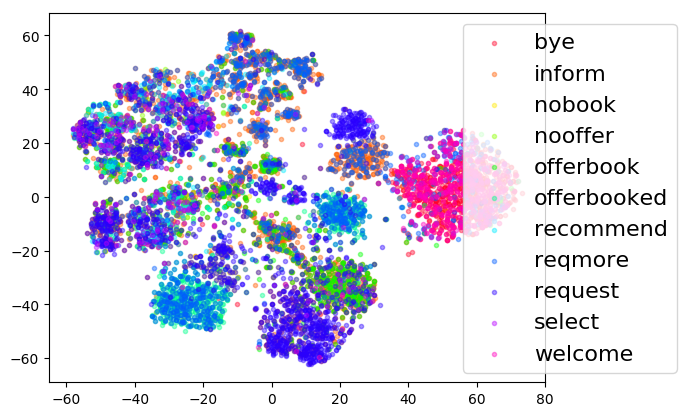}
		\caption{LiteAttnCat*, action}
	\end{subfigure}%
	\begin{subfigure}{.25\textwidth}
		\centering
		\includegraphics[width=\linewidth]{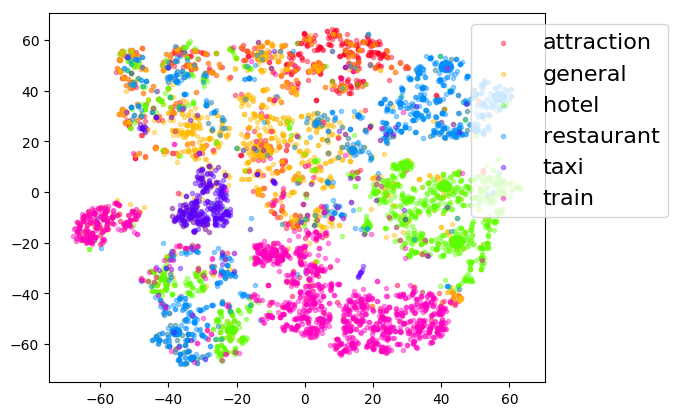}
		\caption{LiteAttnCat+RL*, domain}
	\end{subfigure}%
	\begin{subfigure}{.25\textwidth}
		\centering
		\includegraphics[width=\linewidth]{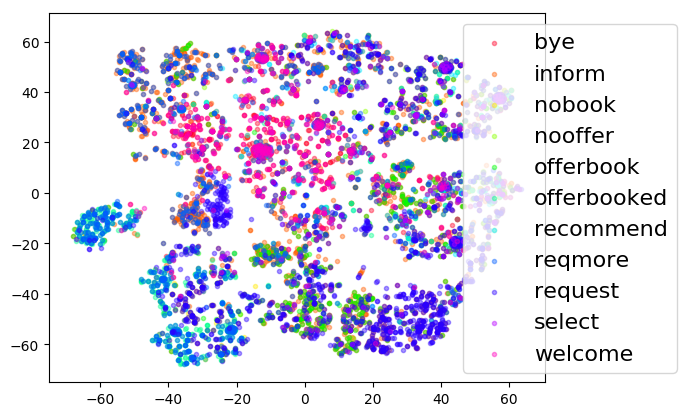}
		\caption{LiteAttnCat+RL*, action}
	\end{subfigure}
	
	\caption{LiteAttnCat* and LAVA\_kl\_cat latent space projection before and after RL. While LiteAttnCat loses action cluster definition, our LAVA\_kl\_cat improves both domain and action clusters definition without causing significant latent space transformation. Higher resolution version is in Appendix~\ref{app:plot}.}
	\label{fig:plot}

\end{figure}

We visually assess the latent space by projecting the latent action of each input in the training set with t-SNE~\cite{maaten2008visualizing} and analyzing the cluster that formed w.r.t. domain and action labels. We compare our best proposed model LAVA\_kl\_cat with the baseline LiteAttnCat, before and after RL, presented in Figure \ref{fig:plot}. While LiteAttnCat sees good domain cluster definition after SL and RL, it loses some action cluster definition after RL. On the other hand, when training on the informed prior, we obtain clusters that are tighter and farther apart from each other. Performing RL on top of SL moves the clusters inwards and improves cluster definition without causing significant transformation of the latent space. This indicates that the proposed method allows the model to put more focus on learning a dialogue policy without having to radically modify the latent action representation. This also signals that the model is equipped with an effective action space since the beginning of RL, which boosts learning.


\subsection{Latent Variable Traversal}
Latent variable traversal can be employed to qualitatively analyze the relationship between different parts of the latent space. This is done by taking two points from the latent space, traversing the space between them by interpolating intermediate variables, and generating a sample for each variable. A latent space that meaningfully encode the generative factors would be able to produce reasonable samples with gradating similarities to the opposite ends of the traversal.

We select two dialogue contexts where the target responses perform similar actions but in different domains. Traversal of latent variables from LAVA\_kl\_cat and reproduced LiteAttnCat models for these contexts is presented in Table \ref{tab:traversal}. We observe that the proposed model is able to transition smoothly from an inform and offerbook action in the hotel domain, to providing booking confirmation in the taxi domain. On the other hand, latent representations trained without the support of VAE generate other actions in the traversal, signaling that the action features are not encoded effectively in the latent representation. It is also worthwhile to note that the two actions are closer to each other in the LAVA\_kl\_cat model, and farther apart in the LiteAttnCat model. With LAVA\_kl\_cat, traversal after RL yields the same responses while LiteAttnCat shows improved traversal. This echoes our previous analysis that the proposed method is equipped with an action-characterized action space since the beginning of RL, which supports effective and practical RL with end-to-end dialogue models.


\begin{table}
    \small
    \centering
    \begin{tabular}{|p{0.97\linewidth}|}
    \hline
    LiteAttnCat* \\
    \hline
        \textbf{yes, {[hotel\_name]} is a {[value\_pricerange]} -ly priced {[value\_count]} star guesthouse located in the {[value\_area]}. it is {[value\_pricerange]} -ly priced and has {[value\_count]} stars . would you like to book a room?} \\
         \hspace{0.1cm} sure, i can help you with that . what would you like to know?\\
         \hspace{0.1cm} i have the {[hotel\_name]} located at {[hotel\_address]}. would you like me to book it for you?\\
         \hspace{0.1cm} i have the {[restaurant\_name]} located at {[restaurant\_address]}. would you like to book a table?\\
         \hspace{0.1cm} the address is {[restaurant\_address]} and the phone number is {[restaurant\_phone]}.\\
         \hspace{0.1cm} the {[hotel\_name]} is a {[value\_count]} star guesthouse in the {[value\_area]} area . it is {[value\_pricerange]} -ly priced and has free wifi and parking . would you like me to book it for you?\\
         \textbf{i have booked you a taxi . it will be a {[taxi\_type]} and the contact number is {[taxi\_phone]}.}\\ 
         \hline 
         LAVA\_kl\_cat\\
         \hline
         \textbf{[hotel\_name] is a guesthouse in the [value\_area] area . it is [value\_pricerange] -ly priced and has [value\_count] stars . would you like to book a room?}\\
         \hspace{0.1cm} i would recommend [hotel\_name]. it s a [value\_count] star guesthouse in the [value\_area]. would you like to book a room?\\
         \hspace{0.1cm} the reference number for the train is [train\_reference].\\
         \hspace{0.1cm} i am sorry, i am not able to book that . i can book you a room at the [restaurant\_name] if you would like.\\
         \hspace{0.1cm} i am sorry, but i am unable to book it right now . is there anything else i can help you with?\\
         \hspace{0.1cm} i can book that for you now.\\
         \textbf{i have booked you a [taxi\_type]. the contact number is [taxi\_phone]. can i help you with anything else?}\\
         \hline

    \end{tabular}
    \caption{Latent variable traversal between two responses with dialogue actions related to booking in different domains. Traversal on LAVA\_kl\_cat shows smooth transition with consistent underlying action.}
    \label{tab:traversal}
    \vspace{-0.5cm}
\end{table}

\section{Conclusion and Future Work}

This work acts as proof of concept that we can induce action-characterized latent representations in an unsupervised manner to facilitate a more practical and effective RL with end-to-end dialogue models. We explore ways to obtain action-characterized latent representations via response variational auto-encoding, which captures generative aspects of responses. Treating these latent representations as actions allows effective optimization with RL. Unlike contemporary transformer-based approaches, our method requires no additional data and has low computational cost. We are able to achieve state-of-the-art success rate on the challenging MultiWoZ 2.0 corpus on both context-to-response generation as well as the end-to-end modeling task. Our analyses show that the proposed methods result in latent representations that cluster well w.r.t. domain and action labels, and encode similar actions close to each other. In this paper, we utilize simple recurrent models to highlight the merit of the proposed training methods, which means each component can be replaced with stronger models to further improve performance. We believe our method has high potential for end-to-end domain adaptation and offline policy learning with RL. We look forward to improve our work by utilizing longer context and performing RL in a stricter setting where the dialogue trajectory is more accurately estimated. We would also like to conduct human evaluation and analyze how our model performs in real dialogue interaction. 

\section*{Acknowledgements}
N. Lubis, M. Heck, and C. van Niekerk are supported by funding provided by the Alexander von Humboldt Foundation in the framework of the Sofja Kovalevskaja Award endowed by the Federal Ministry of Education and Research, while C. Geishauser, H-C. Lin and M. Moresi are supported by funds from the European Research Council (ERC) provided under the Horizon 2020 research and innovation programme (Grant agreement No. STG2018\_804636). Computing resources were provided by Google Cloud.


\bibliographystyle{coling}
\bibliography{refs.bib}

\appendix

\section*{Appendix}

\section{Additional Results on MultiWoZ 2.1}
\label{app:results}
For a more complete comparison with existing works, we report additional results of our best model. We trained and tested LAVA\_kl\_cat + RL with MultiWoZ 2.1 dataset~\cite{eric2019multiwoz}, and computed the match and success rates of the model using a new evaluation script published in the official MultiWoZ repository\footnote{\texttt{https://github.com/budzianowski/multiwoz}}. MultiWoZ 2.1 includes corrections to the dialogue state labels and canonicalization of the slot values in the utterance, reducing the labeling error and typos introduced by human worker. As previously explained, the new evaluation script differs to the old one in the treatment of one specific case in the train domain, where a train matching user requirement is found but the user does not request the train ID. The old script underestimates the model performance as the train ID is checked regardless, while the new evaluation script do not check this further. The results are presented in Table~\ref{tab:additional_results}. Training and testing with MultiWoZ 2.1 yields lower match and success rates, however BLEU score is improved. At the time of publication, our best model achieves state-of-the-art results.

\begin{table}[h]
\small
    \centering 
    \begin{tabular}{|l|c|c|c||c|c|}
        \hline
        Model              & Match & Success & BLEU & Transformer & RL\\ \hline
        SimpleTOD \cite{hosseini2020simple} & 85.10 & 73.50 & 0.16 & $\checkmark$ & -\\ \hline
        MarCo \cite{wang2020multi} & 92.50 & 77.80 & 0.19 & $\checkmark$ & -\\ \hline
        HDNO \cite{wang2020modelling} & 92.80 & 83.00 & 0.18 & - & $\checkmark$ \\ \hline \hline
        LAVA\_kl\_cat + RL (ours) & \textbf{96.39} & \textbf{83.57} & 0.14 & - & $\checkmark$ \\ \hline 
    \end{tabular}
    \caption{Comparison of our best performing model with existing works on MultiWoZ 2.1 data. We also note whether the approaches employ a transformer-based architecture and RL.}
\label{tab:additional_results}
\end{table}

\section{Cluster Overfitting and Its Relation to Calinski-Harabasz Index}
\label{app:overfit}
\begin{figure}[h]
\centering
	\begin{subfigure}{.5\textwidth}
  		\centering
  		\includegraphics[width=\linewidth]{figs/2020-05-12-14-51-49-actz_cat/rl-2020-05-18-10-50-48/rl_y_tsne_sub_domain_tsne_50.png}
  		\caption{LAVA\_kl\_cat + RL, domain. CH: 158.00}
  	\end{subfigure}%
  	\begin{subfigure}{.5\textwidth}
  		\centering
  		\includegraphics[width=\linewidth]{figs/2020-05-12-14-51-49-actz_cat/rl-2020-05-18-10-50-48/rl_y_tsne_sub_act_tsne_50.png}
  		\caption{LAVA\_kl\_cat + RL, action. CH: 41.75}
	\end{subfigure}
	\begin{subfigure}{.5\textwidth}
		\centering
		\includegraphics[width=\linewidth]{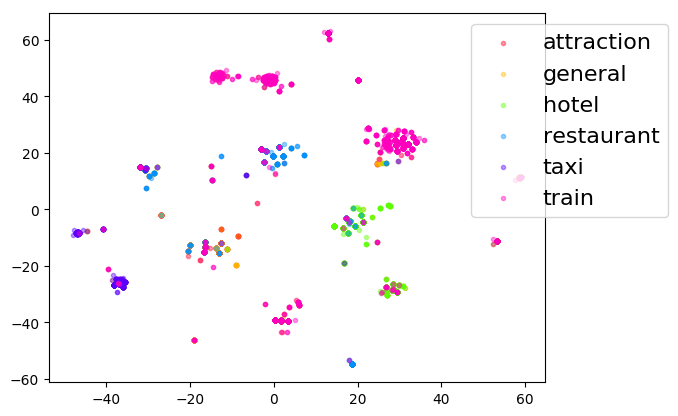}
		\caption{LAVA\_mt\_cat + RL, domain. CH: 247.92}
	\end{subfigure}%
	\begin{subfigure}{.5\textwidth}
		\centering
		\includegraphics[width=\linewidth]{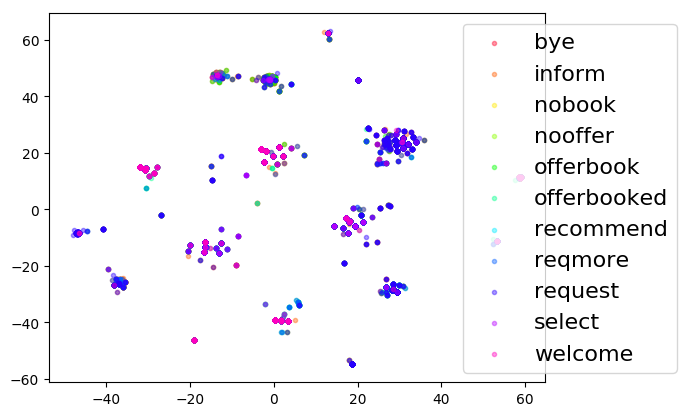}
		\caption{LAVA\_mt\_cat + RL, action. CH: 16.85}
	\end{subfigure}
	\caption{LAVA\_kl\_cat and LAVA\_mt\_cat latent space projection after RL and their resprective Calinski-Harabasz scores (CH). All plots contain the same number of data points. Plots (c) and (d) show that LAVA\_cat\_mt + RL groups the datapoints into fewer clusters, and while this yields high domain CH score, the score for action clusters is low. On the other hand, although the domain CH is lower, LAVA\_kl\_cat shows better clustering fit for both domain and action.}
	\label{fig:overfit_appendix}
	\end{figure}

For a dataset $E$ of size $n_E$ and a center $c_E$ forming $k$ clusters, with each cluster $q$ of size $n_q$ consisting of a centroid $c_q$ and a set of points $C_q$, the Calinski-Harabasz index $CH$ is computed as

\begin{equation}
CH = \frac{tr(B_k)}{tr(W_k)} \times \frac{n_E-k}{k-1},
\end{equation}

\noindent where

\begin{equation}
W_k = \sum_{q=1}^{k} \sum_{x \in C_q}(x-c_q)(x-c_q)^T,
\end{equation}

\begin{equation}
B_k = \sum_{q=1}^{k}n_q(c_q-c_E)(c_q-c_E)^T,
\end{equation}

\noindent and $tr(B_k)$ and $tr(W_k)$ are the traces of the between-cluster dispersion and the inter-cluster dispersion, respectively. It is evident that the smaller the inter-cluster dispersion, the higher the score will be. However in our experiements we find that high score could be a sign of cluster overfitting in the latent space, that is when the clusters do not preserve enough variability of the target responses. For example, when each domain is grouped into one small cluster and each cluster is placed far from each other, scoring on domain labels may translate to extremely high values, yet this is not desired in practice as we would like the latent representation to capture within-cluster varieties as well, or maybe split one domain into several clusters to better distinguish other action-relevant information. Furthermore, overfitting w.r.t. domain label set typically means poor fit w.r.t. action labels. Figure~\ref{fig:overfit_appendix} presents an example.

\section{High Resolution Plots}
\begin{figure}[ht]
	\begin{subfigure}{.5\textwidth}
		\centering
		\includegraphics[width=\linewidth]{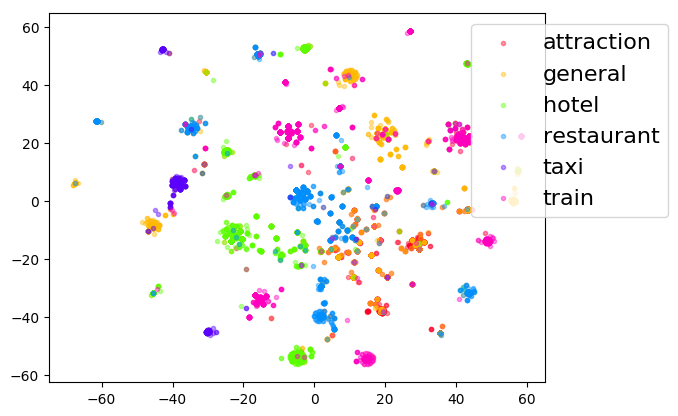}
		\caption{LAVA\_kl\_cat, domain}
	\end{subfigure}%
	\begin{subfigure}{.5\textwidth}
		\centering
		\includegraphics[width=\linewidth]{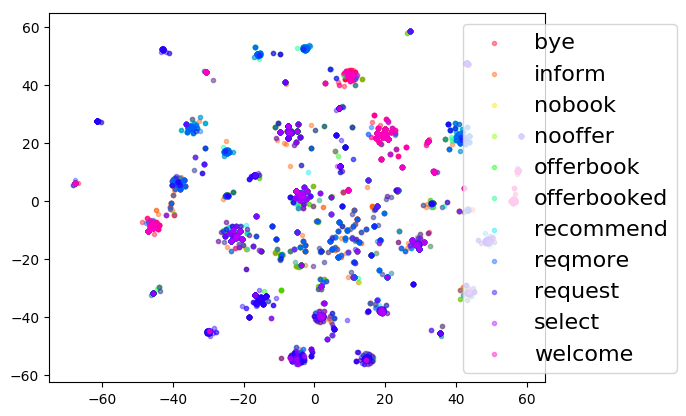}
		\caption{LAVA\_kl\_cat, action}
	\end{subfigure}
	\begin{subfigure}{.5\textwidth}
  		\centering
  		\includegraphics[width=\linewidth]{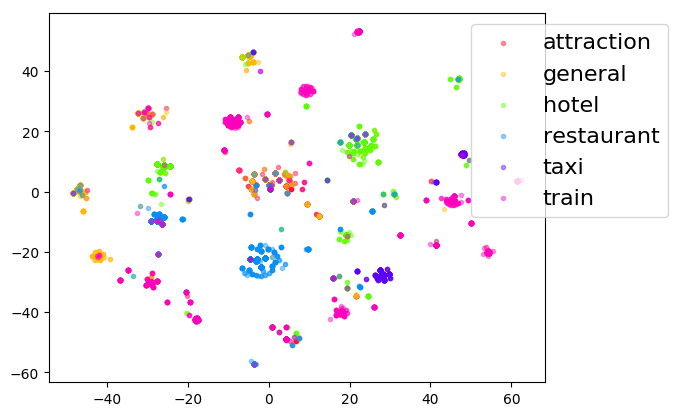}
  		\caption{LAVA\_kl\_cat + RL, domain}
  	\end{subfigure}%
  	\begin{subfigure}{.5\textwidth}
  		\centering
  		\includegraphics[width=\linewidth]{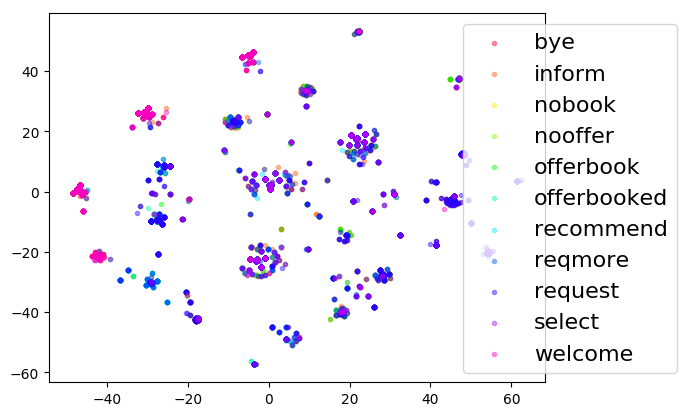}
  		\caption{LAVA\_kl\_cat + RL, action}
	\end{subfigure}
	\begin{subfigure}{.5\textwidth}
		\centering
		\includegraphics[width=\linewidth]{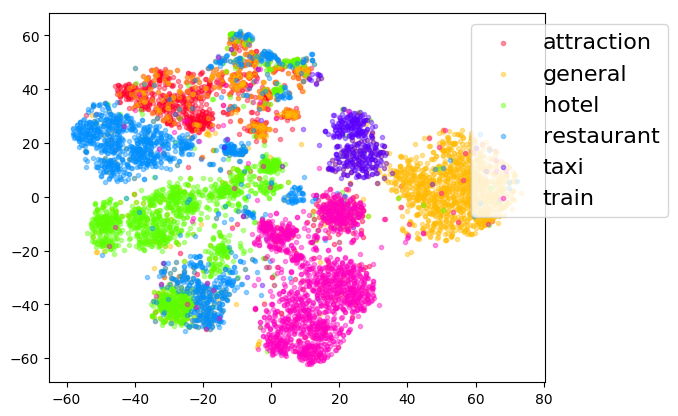}
		\caption{LiteAttnCat*, domain}
	\end{subfigure}%
	\begin{subfigure}{.5\textwidth}
		\centering
		\includegraphics[width=\linewidth]{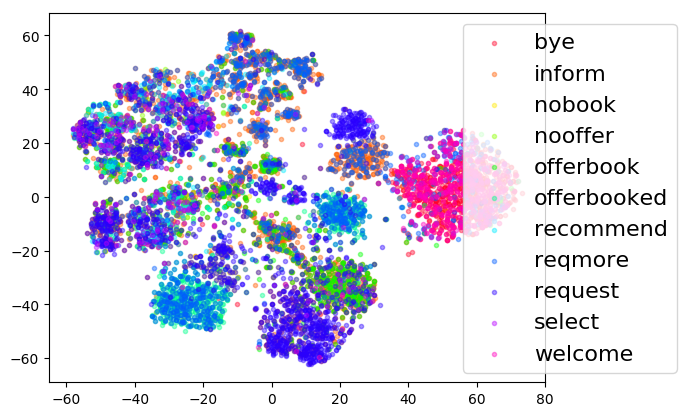}
		\caption{LiteAttnCat*, action}
	\end{subfigure}
	\begin{subfigure}{.5\textwidth}
		\centering
		\includegraphics[width=\linewidth]{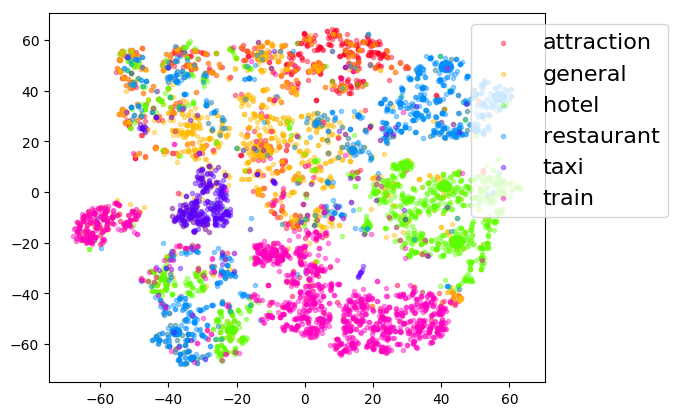}
		\caption{LiteAttnCat + RL*, domain}
	\end{subfigure}%
	\begin{subfigure}{.5\textwidth}
		\centering
		\includegraphics[width=\linewidth]{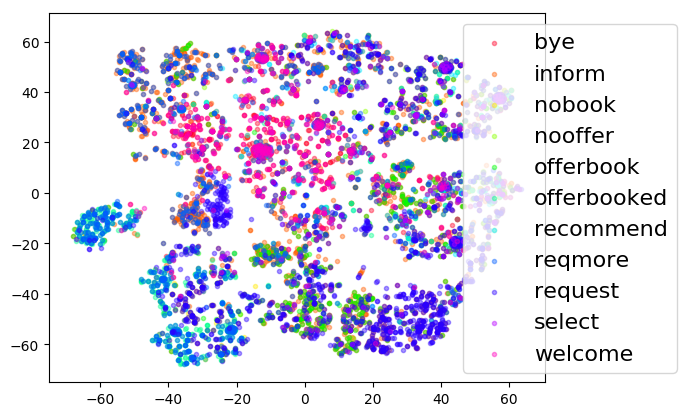}
		\caption{LiteAttnCat + RL*, action}
	\end{subfigure}
	
	\caption{LiteAttnCat* and LAVA\_kl\_cat latent space projection before and after RL. While LiteAttnCat loses action cluster definition, our LAVA\_kl\_cat improves both domain and action clusters definition without causing significant transformation of the latent space.}
	\label{fig:app_plot}
\end{figure}

We reproduce the cluster projections from Figure~\ref{fig:plot} in high resolution, presented in Figure~\ref{fig:app_plot}.
\label{app:plot}

\end{document}